\documentclass[10pt,twocolumn,letterpaper]{article}

\usepackage{cvpr}              %
\usepackage[dvipsnames]{xcolor}

\usepackage{indentfirst}

\usepackage{tabularx}

\definecolor{cvprblue}{rgb}{0.21,0.49,0.74}
\usepackage[pagebackref,breaklinks,colorlinks,citecolor=cvprblue]{hyperref}

\def\paperID{16594} %
\def\confName{CVPR}
\def\confYear{2024}

\title{Volumetric Cloud Field Reconstruction}

\author{Jacob Lin$^1$ \quad Miguel Farinha$^2$ \quad Edward Gryspeerdt$^1$ \quad Ronald Clark$^2$ \\
$^1$Imperial College London \qquad $^2$University of Oxford\\
{\tt\small linjacob2@gmail.com, \{miguel.farinha, ronald.clark\}@cs.ox.ac.uk, e.gryspeerdt@imperial.ac.uk}
}

\begin{document}
\maketitle

\begin{abstract}
Volumetric phenomena, such as clouds and fog, present a significant challenge for 3D reconstruction systems due to their translucent nature and their complex interactions with light. Conventional techniques for reconstructing scattering volumes rely on controlled setups, limiting practical applications. This paper introduces an approach to reconstructing volumes from a few input stereo pairs. We propose a novel deep learning framework that integrates a deep stereo model with a 3D Convolutional Neural Network (3D CNN) and an advection module, capable of capturing the shape and dynamics of volumes. The stereo depths are used to carve empty space around volumes, providing the 3D CNN with a prior for coping with the lack of input views. Refining our output, the advection module leverages the temporal evolution of the medium, providing a mechanism to infer motion and improve temporal consistency. The efficacy of our system is demonstrated through its ability to estimate density and velocity fields of large-scale volumes, in this case, clouds, from a sparse set of stereo image pairs.
\end{abstract}    
\section{Introduction}
\label{sec:intro}

Volumetric phenomena, such as clouds, smoke, steam, and fog, play an integral role in shaping the natural world and our perception of it. Whether limiting visibility, driving severe weather, or providing picturesque backdrops, volumetric phenomena such as clouds have a significant visual, practical, and cultural importance in our daily lives, as well as driving uncertainties in future climate projections~\cite{forster22}. 

\begin{figure}[h!]
    \centering
    \includegraphics[width=1\linewidth]{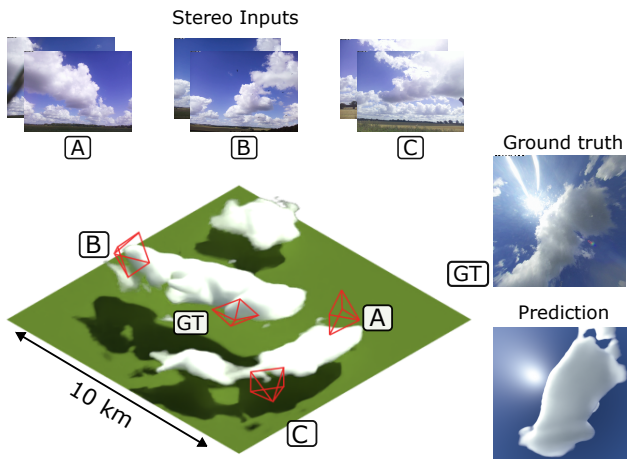}
    \caption{Our method reconstructs volumetric density fields from sparse stereo views. Our stereo depth carving module provides a coarse volumetric estimate, allowing our model to recover volume shapes even with limited views. Predicting per-frame density estimates, we additionally use an advection module that leverages the smooth motion of volumes to refine our reconstructions.}
    \label{fig:banner}
\end{figure}

This has led to an increasing interest in the computer vision and graphics communities in retrieving the density field of a volume from a set of RGB images~\cite{in-situ-view, material_dicts, imaging_scattering, chu2022physics, franz2021global}. The majority of existing works have focused on computed-tomography or inverse-rendering type approaches with highly controlled setups or synthetic data for experiments~\cite{levis2015airborne, levis2017multiple, levis2020multi, aides2020distributed, veikherman2015clouds, tzabari2022settings, gkioulekas2016evaluation, in-situ-view}. In these controlled scenarios, the image formation process can be systematically modeled, enabling the application of inverse methods to estimate the properties of the medium under observation. 

One important application of the retrieval of volumes in this area is the remote sensing of the atmosphere. With their central role in the water cycle and energy budget, the measurement of the spatial distribution of airborne particles, such as water droplets and ice crystals, that makeup clouds has been an important objective. To this extent, recent works have shown that it is possible to use deep learning based models to make the estimation more robust, allowing the structure of clouds to be recovered from satellite observations~\cite{3deepct, vip_ct}. However, while these works have made significant strides in enabling real-world retrieval of scattering volumes, they still require an expensive setup, i.e. unobstructed orthographic views of the top of the volume, limiting the reach of their application.

Furthermore, there are a number of significant computer vision challenges associated with reconstructing volumetric phenomena. Firstly, most volumes, like clouds have a uniquely complex visual appearance characterized by little texture detail or stable features. The reason for this is that clouds exhibit different types of scattering: their inner core predominantly exhibits multi-scattering and therefore has a diffuse appearance, while the boundary primarily exhibits single-scattering leading to significant changes in appearance with viewpoint~\cite{forster21veiled}. Furthermore, real-world settings introduce additional complications, such as glare from the sun, which pose significant modeling challenges. Spanning multiple kilometers, with views often obscured by other neighboring clouds, obtaining views of a cloud from all directions is difficult, particularly when considering the logistical complexities associated with deploying cameras in the field as part of measurement programs. 

Our proposed method aims to overcome these challenges by using a stereo module in combination with a 3D CNN for predicting the spatially varying properties of volumes from a sparse set of stereo pairs. Our stereo module gives an initial estimate of volume shapes by carving empty space, establishing coarse volume boundaries. When embedded in a large-scale flow, the volumes will usually exhibit smooth motion and therefore we propose an advection module to make use of the temporal characteristics inherent to the data. Specifically, our approach estimates a velocity field, ensuring enhanced consistency in our reconstructions.

To summarize, our contributions are: \\
1) We propose a stereo depth carving module that gives our 3D CNN a coarse volumetric estimate, allowing our method to reconstruct volumetric fields even with a sparse set of input views. \\
2) We leverage the temporal dynamics of volumes, modeling the motion of the medium and improving volumetric reconstruction through our advection module. \\
3) We release two cloud datasets for stereo few-view cloud field reconstruction; a synthetic dataset for training, and a real-world dataset for evaluation. Using these datasets, we demonstrate that our method is able to recover accurate cloud shapes and cloud positions using a sparse set of stereo views. \\

\section{Related Work}
\label{sec:related}

\begin{figure*}[h!]
    \centering
    \includegraphics[width=1\linewidth]{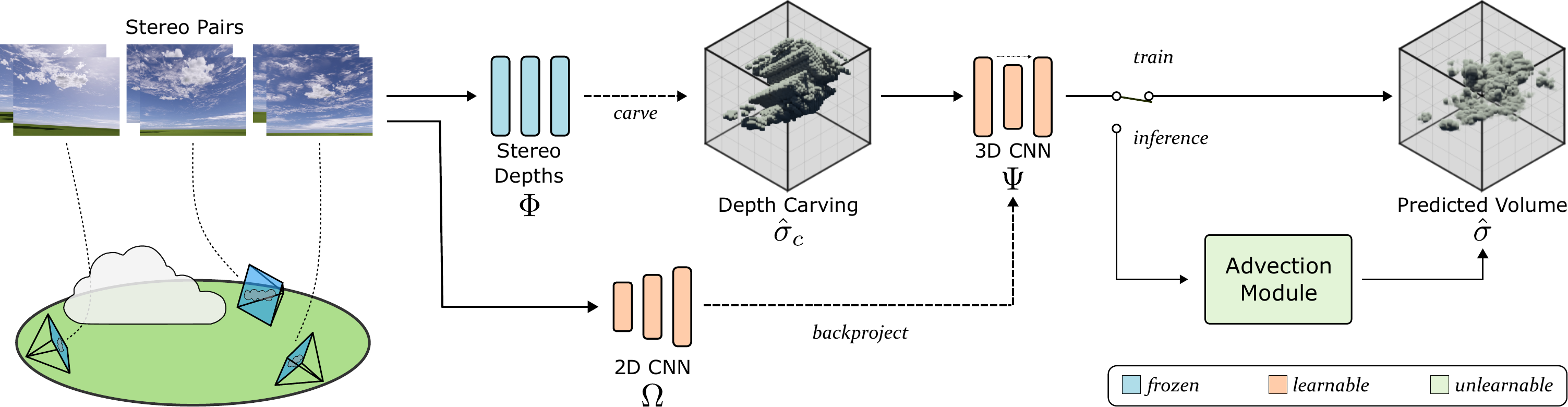}
    \caption{\textbf{Model overview.} Our model reconstructs volumetric fields using a sparse set of stereo images. We address the lack of views by using depths from a pre-trained stereo predictor to carve empty space around volumes, creating an initial coarse estimate of the density field $\hat{\sigma}_c$. Image features and a 3D CNN volumetric shape prior are both trained on synthetic data, such that the 3D CNN $\Psi$ learns to use image features to refine the initial coarse density estimate. At inference time, our advection module uses the motion of volumes to integrate densities from multiple timesteps, outputting a final density field $\hat{\sigma}$.}
    \label{fig:method_overview}
\end{figure*}

\paragraph{Physics-based volume reconstruction.} The advent of deep learning has provided the computer vision community with new tools to address the reconstruction of volumetric phenomena, such as smoke~\cite{kim2022deep, chu2022physics, eckert2019scalarflow, kim2019transport}, dynamic fluids~\cite{thapa2020dynamic, franz2021global}, water~\cite{levy2023seathru} and clouds~\cite{liu2020hyperspectral, zhang2018cloudnet, zheng2021neural, 3deepct, vip_ct}. Notably, 3DeepCT~\cite{3deepct} performs cloud reconstruction using a fully-convolutional architecture trained on multi-view physical cloud simulations, achieving reconstruction quality comparable to the one obtained by traditional explicit physics-based methods~\cite{levis2015airborne, levis2017multiple, levis2020multi} with five orders of magnitude faster inference time. However, this method couples the resolution of the reconstruction to the resolution of the input images and requires using a 2D CNN to implicitly learn camera geometries, leading to limited capabilities and low flexibility. To alleviate this, VIP-CT~\cite{vip_ct} proposes an architecture comprising a decoder that takes as input, image features extracted using a CNN and a set of vectors that encode the camera and coordinate frame. A key difference between VIP-CT and our method is that we learn a 3D prior rather than separately querying each 3D location in a voxel grid for feature extraction. Furthermore, both 3DeepCT and VIP-CT are dependent on silhouette-based space carving~\cite{spacecarving}, which requires both many views and good segmentation masks. In contrast, instead of silhouette-based space carving, our method utilizes depth carving (space carving with depth maps) which is better able to estimate free-space and non-convex shapes.

\paragraph{Neural radiance fields.} Reconstruction of volumes from RGB images has been a long-standing problem in computer vision and graphics. One of the most promising research directions for volumetric reconstruction is neural scene representations, which implicitly model scenes as the weights of neural networks~\cite{neural_volumes, meshry2019neural, sitzmann2019deepvoxels, thies2019deferred, nerf}. Notably, Neural Radiance Fields (NeRF)~\cite{nerf} have shown high-quality 3D reconstruction of complex real-world scenes by using a Multilayer Perceptron (MLP) to implicitly encode volumetric density and color. However, NeRF requires a very long per-scene training process to obtain high-quality reconstructions, making this method infeasible for many applications. Subsequent works aimed to reduce the substantial per-scene optimization time~\cite{yu2021plenoctrees, garbin2021fastnerf, hu2022efficientnerf, piala2021terminerf, liu2020neural}. For instance, DVGO~\cite{sun2022dvgo} replace NeRF's implicit MLP representation with a dense voxel grid to directly model the 3D geometry, while Instant-NGP~\cite{muller2022instantngp} propose to jointly train a multi-resolution hash encoding of feature vectors with the NeRF model MLPs. Despite greatly improving NeRF model training and inference speed, these approaches require many posed views of a scene for optimization. This is in contrast to our method which only requires a sparse set of input views for reconstructing a given volume.  

\paragraph{Few-view 3D reconstruction.} Recently, numerous methods have been presented for scene reconstruction using only a sparse set of input images~\cite{yu2021pixelnerf, chen2021mvsnerf, lin2023visionnerf, gu2023nerfdiff, wang2021ibrnet, kanaoka2023manifoldnerf, xu2022sinnerf}. PixelNeRF~\cite{yu2021pixelnerf} overcomes NeRF's inability to share knowledge between scenes by using the pre-trained layers of a CNN to extract image features that are used as a prior to condition NeRF. MVSNeRF~\cite{chen2021mvsnerf} leverages deep multi-view stereo (MVS) techniques by using a 3D CNN to reconstruct a neural scene encoding volume which facilitates the generalization to unseen testing scenes. IBRNet~\cite{wang2021ibrnet} learns a generic view interpolation network comprising an MLP and ray transformer to obtain colors and densities by aggregating information present in a sparse set of nearby views. These methods require nearby input views and are mainly focused on view interpolation. Unlike these methods, our method is capable of performing high-quality reconstructions of whole volumes with severe occlusions and glare from the sun while naturally dealing with distant and large input view changes.

\paragraph{Stereo depth estimation.} Given two RGB images, stereo depth estimation methods aim to estimate a depth map by learning how to match pixels across the input images along rectified (horizontal) epipolar lines. RAFT-Stereo~\cite{lipson2021raftstereo} extends the RAFT~\cite{teed2020raft} architecture for optical flow by introducing a 3D lightweight cost volume and using multi-level 2D CNNs, that allow efficiently passing information across the image, to process the stereo cost volume. More recently, GMStereo~\cite{xu2023gmstereo} propose a unified framework for different matching tasks, such as flow, disparity, and depth. Given its robustness to challenges like occlusions, we use as our stereo depth predictor, GMStereo, which has additionally been finetuned on synthetic cloud data. In the context of 4D cloud reconstruction, COGS~\cite{cogs} uses stereo reconstruction from three pairs of stereo cameras to reconstruct cloud volumes at different time intervals.

\section{Method}
\label{sec:method}
In this section, we present our approach for reconstructing volumes using a sparse set of stereo views. Namely, given an input of $M$ posed stereo pairs $\{I_{i, L}\}$, $\{I_{i, R}\}$, grouped together as $\{I_i\}$, together with their corresponding camera poses $\{P_i\}$ and intrinsics $\{K_i\}$, we produce an estimate $\hat{\sigma} \in \mathbb{R}^{N_x \times N_y \times N_z}$ of a volumetric density field $\sigma$. 

\subsection{Model Overview}
Our model (shown in Figure~\ref{fig:method_overview}) uses a pre-trained stereo depth estimator, $\Phi$, to produce a set of M depth maps $\{\hat{D}_i\}$. Using the camera parameters and the depth maps, our depth carving module (described in Section~\ref{sec:depth_carving}) produces an initial coarse density grid $\hat{\sigma}_c$ from a sparse set of views. Image features $\{F_i\}$ are extracted from the images using a 2D CNN, $\Omega$. These are backprojected using the corresponding camera poses and then averaged across the different views to form a feature volume $V$. Using the coarse density grid $\hat{\sigma}_c$ and the feature volume $V$, a 3D CNN, $\Psi$, then predicts the final density estimate $\hat{\sigma} = \Psi(\hat{\sigma}_c, V)$.

Lastly, at inference, our method models the dynamics of the volume through an advection module (described in Section~\ref{sec:wind_motion}) such that density fields across multiple timesteps can be integrated to output a single refined volume. 

\begin{figure}[h!]
    \centering
    \includegraphics[width=1\linewidth]{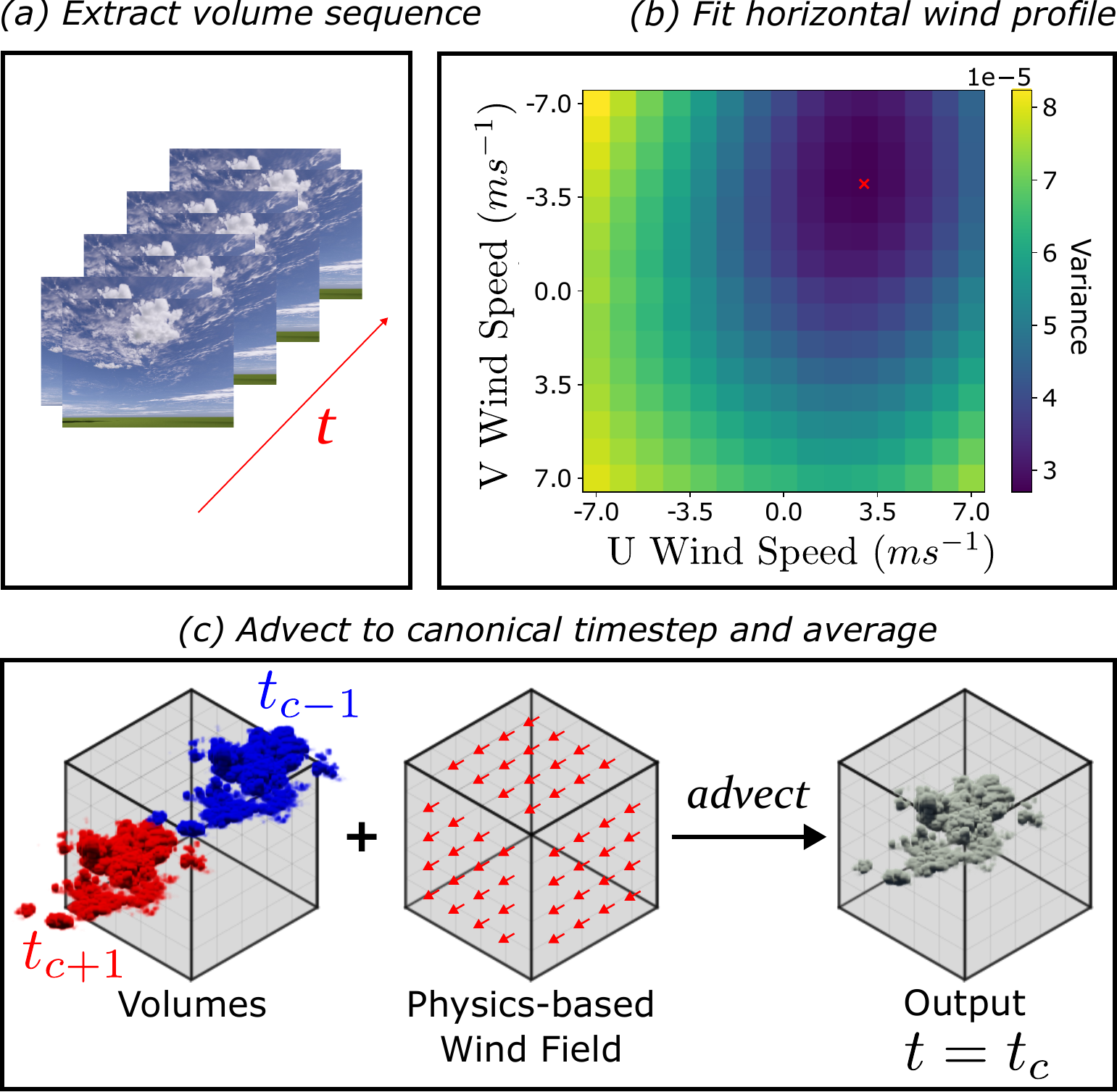}
    \caption{\textbf{Overview of our advection module.}  (a) Given a temporal sequence of multi-view stereo pair inputs, we predict a sequence of density fields. (b) A constrained physics-based wind field is then fit such that the motion of the density fields is modeled. (c) Using the wind field, cloud fields over multiple timesteps are then integrated into a single density estimate that is temporally consistent, with less noise and better volumetric shapes.}
    \label{fig:advection}
\end{figure}

\subsection{Stereo Depth Carving}
\label{sec:depth_carving}
Given the sparse input views, our depth carving module is essential to provide our model with an initial prior on volume shape and location.

Let $d(\mathbf{x}, \hat{D}_i, P_i, K_i)$ be the signed distance at a point $\mathbf{x} \in \mathbb{R}^3$, that is obtained by backprojecting a single depth map $\hat{D}_i$ using the corresponding camera parameters $P_i$ and $K_i$. Then from the M depth maps, we produce a set of M carvings \{$C_i \in \mathbb{R}^{N_x \times N_y \times N_z}$\} by considering, for each depth map, the following carving rule at each grid point $\mathbf{x}$: 
\begin{equation}
    \texttt{carve} \rightarrow
    \begin{cases}
      1, & \text{if $d(\mathbf{x}, \hat{D}_i, P_i, K_i) < \epsilon$} \\
      0, & \text{otherwise}
    \end{cases}
\end{equation}
where $\epsilon$ is a margin to allow for errors in the estimated depths. In practice we set $\epsilon = 1000 \, \text{m}$. 

\newpage
Our coarse volume estimate $\hat{\sigma}_c$ is then obtained by combining the set of carvings:
\begin{equation}
    \hat{\sigma}_{c} = C_0 \odot C_1 \odot \dots \odot C_M 
\end{equation}
where $\odot$ denotes an element-wise product across the grid. In other words, our method uses stereo depths to carve out empty space in front of volumetric surfaces and along rays that do not intersect any volumes (here we consider the sky to be at an infinite depth).

\paragraph{Comparison to existing approaches.} Our depth carving improves over silhouette-based space carving~\cite{spacecarving} in that ours is not limited to a convex hull and is capable of carving empty space in front of volumes as well. This can have a significant effect in real-world settings, for example with clouds, where distant clouds along the horizon or high-altitude cloud layers would pose a significant issue to segmentation-based space carving. Additionally, previous approaches, such as VIP-CT~\cite{vip_ct} and 3DeepCT~\cite{3deepct}, have only used space carving as a mask applied to the output. Our method instead uses our depth carving in a learned manner by further refining the initial coarse estimate with a 3D CNN.

\subsection{Physics-Based Wind Motion}
\label{sec:wind_motion}

We model the dynamics of reconstructed volumes to further refine density predictions at inference time.

For clouds, large-scale motion is dominated by horizontal winds that mainly change as a function of altitude. However, as our method reconstructs volumes in a height-restricted domain, our cloud fields are generally limited to single low-level cumulus cloud layers that are vertically thin. The wind field can therefore be modeled using only 2 horizontal scalar components. Our advection module (illustrated in Figure~\ref{fig:advection}) uses this physics-constrained wind field to model the smooth motion of clouds, improving our predictions during inference. 

Given a temporal sequence of frames, we first use our model to extract per-frame densities $\{\hat{\sigma}_t\}$, where for simplicity we let $t = 0, 1, \dots, T - 1$. To integrate these densities we define an advection operator $\texttt{advect}(\hat{\sigma}_t, u, v, t - t_c)$. The operator takes a density field and advects it from time $t$ to time $t_c = \frac{(T - 1)}{2}$, using two scalar horizontal wind components $u$ and $v$.

Using the advection operator, we fit a horizontal wind profile to the sequence by optimizing for:
\begin{equation}
    \hat{u}, \hat{v} = \mathop{\mathrm{arg\,min}}_{u, v} \overline{\mathrm{Var}}(\texttt{advect}(\hat{\sigma}_i, u, v, i - t_c))
\end{equation}
where $i=0, 1, \dots, T - 1$, so that $\overline{\text{Var}}(\cdot)$ is taking the variance over $T$ advected density fields and then taking the mean over the whole variance grid. In practice, we implement this optimization through a simple grid search. The optimized wind profile can then be used to integrate multiple frames into a single volume at time $t_c$:
\begin{equation}
    \hat{\sigma}_c = \frac{1}{T} \sum_{t=0}^{T-1} \texttt{advect}(\hat{\sigma}_t, \hat{u}, \hat{v}, t - t_c)
\end{equation}
This results in a refined density with better cloud shapes and fewer artifacts, while also naturally having improved temporal consistency. 

\subsection{Training}
We train our model using synthetic cloud data (detailed in Section~\ref{sec:synthetic_data}), such that a 3D CNN-learned cloud shape prior and cloud-specific image features can be trained through ground truth density grids $\sigma$. During training, we use the ground truth to consider cloud voxels and empty voxels separately as $[\sigma_{cloud}$, $\sigma_{empty}]$ for the ground truth, and $[\hat{\sigma}_{cloud}$, $\hat{\sigma}_{empty}]$ for the predictions. With this separation, let $N$ be the number of empty voxels, and $N^c$ be the number of cloud voxels. Our loss function is then given by:
\begin{equation}
    \mathcal{L} = \frac{1}{N^{c}} \| \sigma_{cloud} - \hat{\sigma}_{cloud} \|_1 + \frac{\lambda}{N} \| \sigma_{empty} - \hat{\sigma}_{empty} \|_1
\end{equation}
where lambda is a hyperparameter $\lambda \leq 1$.

As cloud fields are naturally sparse, most density grids will mainly be occupied by empty voxels. By taking the mean of the cloud voxels and the empty voxels separately, our loss effectively weighs the cloud voxels higher relative to the empty voxels. This helps prevent local minima where the model only predicts empty voxels correctly.

During training, we also apply color augmentation to the stereo images, and we randomly exclude up to two stereo pairs as well.

\paragraph{Implementation Details.} 
Our model reconstructs volumes from heights of 400 m to 4000 m, within a 10 km x 10 km area. We choose a voxel size of 50 m, giving us dimensions of $\hat{\sigma} \in \mathbb{R}^{200 \times 200 \times 72}$. We use a simple three-layer 2D CNN which is not pre-trained, while our 3D CNN follows a UNet-like~\cite{unet} architecture. For depth estimation, we use a pre-trained GMStereo~\cite{xu2023gmstereo} which has been finetuned on synthetic cloud scenes. Our model is trained in total for 50k steps, where for the first 10k steps, we linearly increase $\lambda$ from 0 to 1. Training takes 40 hours on a single Nvidia Titan RTX GPU.

\section{Cloud Datasets}
\label{sec:data}
\subsection{Real-World Cloud Dataset}
To evaluate our method, we collect real-world data for few-view reconstruction of clouds using stereo cameras. Our camera setup (seen in Figure~\ref{fig:banner}), consists of three stereo camera pairs positioned in an inwards-looking triangle. The baselines vary from 190 m to 350 m, while the distances between the pairs are between 5000 m to 8000 m. For evaluation, we also include an upwards-looking camera in the middle that has a wide field of view (120\textdegree).

Six one-hour-long sequences of cumulus clouds are captured, where the cameras are synchronized by GPS, such that an image is taken every five seconds. We calibrate the cameras using real-time kinematic positioning, giving a position accurate within a centimeter. Images of stars are then used to optimize for the rotation and focal length. We release the dataset such that it is available for public use. 

\subsection{Synthetic Dataset}
\label{sec:synthetic_data}
To train our model, we create a synthetic cloud dataset consisting of stereo cameras with intrinsic and extrinsic parameters based on our real-world cloud data. We create the 3D cloud volumes in Terragen, an application aimed at creating photorealistic natural scenes. We then use Blender's physically-based rendering engine, Cycles, to render the scenes. To simulate high-level clouds, we add a 2D cloud layer aiming to improve the robustness of our model. The following parameters were varied for each image in the dataset: the cloud fraction, cloud optical density, the altitude of the cloud base, the cloud height, sun azimuth, and zenith. By varying these parameters we ensure that the training set captures enough variability for the model to generalize to real-world scenarios. We sample 1000 different conditions with each of these parameters varied. We release our training dataset for public use.

\begin{figure*}[h!]
    \centering
    \includegraphics[width=1\linewidth]{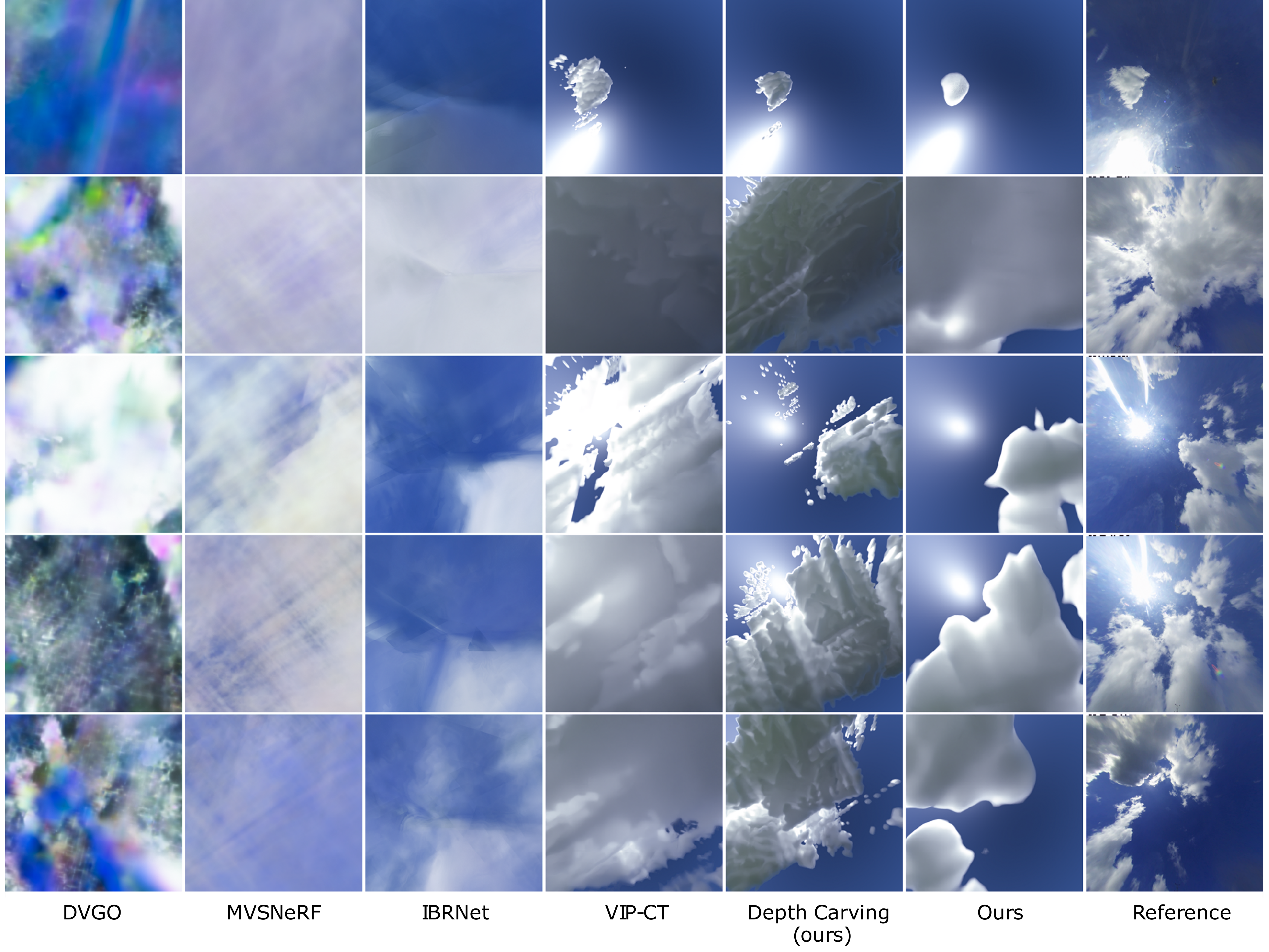}
    \caption{\textbf{Qualitative results.} DVGO~\cite{sun2022dvgo} and MVSNeRF~\cite{chen2021mvsnerf} struggle to recover the volume shape while IBRNet~\cite{wang2021ibrnet}, a view-synthesis method, fares better. VIP-CT~\cite{vip_ct} and 3DeepCT~\cite{3deepct} are able to recover cloud shapes but struggle in more complex cases due to their reliance on silhouette-based space carving. Our method is able to accurately recover cloud shapes.}
    \label{fig:qualitative_results}
\end{figure*}

\begin{figure}[h!]
    \centering
    \includegraphics[width=\linewidth]{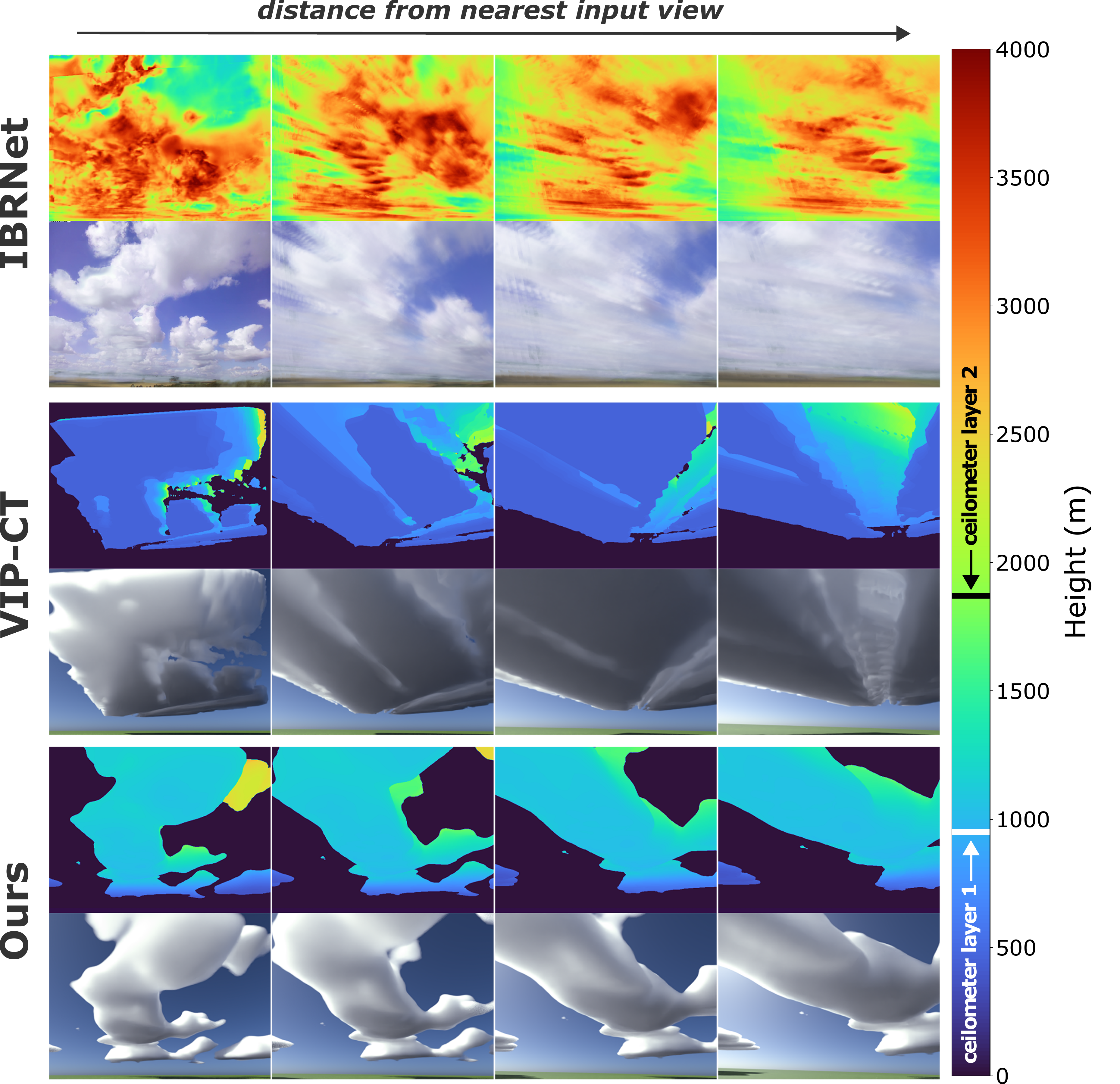}
    \caption{\textbf{Ceilometer cloud height evaluation.} We show height maps and RGB images from viewpoints progressively further away from the nearest input view. Ceilometer data indicates cloud layers at 950 m and 1870 m. IBRNet's~\cite{wang2021ibrnet} inconsistent height maps show poor volumetric shape recovery, degrading with distance from input views. VIP-CT~\cite{vip_ct} underestimates cloud heights, while our method achieves spatially coherent clouds at verified ceilometer heights.}
    \label{fig:interpolated_side}
\end{figure}

\section{Evaluation}
\subsection{Dataset, Baselines, and Netrics}
\paragraph{Dataset.} We conduct experiments using our real-world cloud dataset. For quantitative evaluation, we use the upwards-looking view in the middle of the stereo array, where we manually label segmentation masks for 3 samples from each sequence. In our segmentation masks, we identify and exclude high-altitude clouds that are not visible in the input views. To ensure that our experiments are fair, the samples are systematically selected such that they are evenly spaced out across each sequence.

\paragraph{Baselines.} We compare against three types of methods: cloud-specific reconstruction systems, few-view neural volume rendering approaches with a learned scene prior, and neural volume rendering systems without any learned priors. Specifically, 3DeepCT~\cite{3deepct} and VIP-CT~\cite{vip_ct} are considered for cloud-specific systems, where we train on our synthetic data and perform silhouette-based space carving by thresholding our depths at 20000 m, creating a segmentation mask. For neural volume rendering approaches, we consider MVSNeRF~\cite{chen2021mvsnerf}, IBRNet~\cite{wang2021ibrnet}, and DVGO~\cite{sun2022dvgo}, where the first two have a learned scene prior. For IBRNet we initialize with the released pre-trained model and then train it further on our synthetic data before we lastly do per-scene optimization. The same is done with MVSNeRF, except that the fixed number of source views requires us to only use 5 out of 6 source views, also preventing us from initializing with the pre-trained model.

\paragraph{Metrics.} As cloud and sky appearances are dependent on complex lighting interactions with environmental conditions, we do not use photometric evaluation. We instead evaluate results using segmentation, where we obtain segmentation maps by thresholding opacity maps with a threshold of 0.15. As volumetric rendering methods also reconstruct the sky, we furthermore aim to make results more fair by thresholding the depths at 4000 m, such that far-away points are considered as background. We evaluate results using the standard Jaccard ($\mathcal{J}$) region similarity metric. For this, we exclude clear sky samples as they will always result in zero overlap in segmentation. To ensure we still measure performance in such cases, we also consider the cloud coverage error, which we define as $|\overline{S}_{\!gt} - \overline{S}_{\!pred}|$, where $\overline{S}_{\!gt}$ and $\overline{S}_{\!pred}$ are the mean values of the ground truth and predicted segmentation masks respectively. 

\subsection{Qualitative Results}
To qualitatively evaluate our density fields, we use Blender to render images with a cloud texture applied. 

\paragraph{Sky-view evaluation.} In Figure~\ref{fig:qualitative_results}, we show results using our middle upwards-looking view. We observe that our depth carving by itself is well able to identify empty space, showing its suitability as an initial coarse volumetric prior. Building on this, our 3D CNN-learned cloud shape prior and advection module are able to refine the depth carving, recovering missed cloud densities and further removing empty space. We note in particular that our refined volumes are smooth, with better cloud shapes and positions, while also being less noisy. VIP-CT~\cite{vip_ct} is able to reconstruct clouds in some settings but often fails for more complex cases. DVGO~\cite{sun2022dvgo}, a neural volumetric rendering approach without a learned prior, produces an output with many artifacts. MVSNeRF~\cite{chen2021mvsnerf} and IBRNet~\cite{wang2021ibrnet} on the other hand produce a smooth output but are unable to meaningfully reconstruct clouds. This behavior is not unexpected, as the view used for evaluation is drastically different from the input views.

\paragraph{Ceilometer cloud height evaluation.} Given the logistic difficulties in acquiring additional views for evaluating cloud shapes, we instead consider cloud heights. We obtain cloud height measurements from public ceilometer data taken in the middle of our camera array. Additionally, due to the vertically thin nature of cumulus clouds, we can expect an accurate reconstruction of a cumulus cloud layer to have little variation in height. In Figure~\ref{fig:interpolated_side}, we evaluate 3D cloud shapes by visualizing height maps and rendered RGB images at viewpoints progressively further away from the nearest input view. We observe that IBRNet~\cite{wang2021ibrnet} has noisy height maps and is therefore not recovering coherent cloud shapes. As a result, IBRNet exhibits degrading view interpolation quality as the evaluated view moves away from the nearest input view. VIP-CT~\cite{vip_ct} recovers a volume that is consistent in height, however, due to its reliance on silhouette-based space carving, the carved convex hull underestimates empty space, resulting in too low cloud heights. %
Our method produces spatially coherent cloud shapes, with uniform cloud base heights that match the ceilometer height readings.

\begin{table}[h!]
    \centering
    \begin{tabular}{l c c}
        \toprule
        & $\mathcal{J} \uparrow$ & Coverage Error $\downarrow$ \\
        \midrule
        DVGO~\cite{sun2022dvgo} & 50.4 & 55.1 \\
        \midrule 
        MVSNeRF~\cite{chen2021mvsnerf} & 50.4 & 55.2 \\
        IBRNet~\cite{wang2021ibrnet} & 50.0 & 49.4 \\
        \midrule
        Space Carving~\cite{spacecarving} & 56.5 & 33.9 \\
        3DeepCT~\cite{3deepct} & 57.8 & 33.0 \\
        VIP-CT~\cite{vip_ct} & 57.3 & 33.0 \\
        \midrule
        Depth Carving (ours) & 53.8 & 14.2 \\
        Ours & \textbf{64.1} & \textbf{6.7} \\
        \bottomrule
    \end{tabular}
    \caption{\textbf{Segmentation metrics for our real-world cloud dataset. } Results with silhouette-based space carving and our depth carving are reported as stand-alone components that produce binary density grids.}  %
    \label{tab:quantitative_results}
\end{table}

\begin{figure*}[h!]
    \centering
    \includegraphics[width=\linewidth]{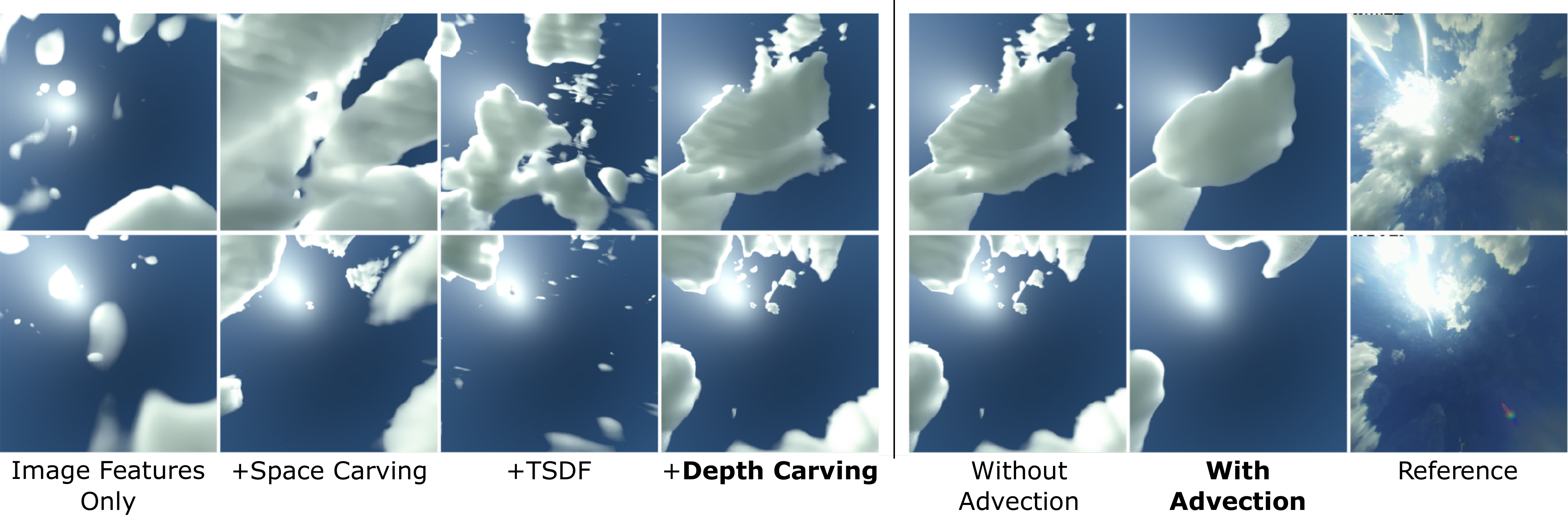}
    \caption{\textbf{Qualitative ablations.} Firstly, we compare different strategies in providing our 3D CNN with an initial estimate of cloud shapes. Secondly, we compare results with and without our advection module. We find that our depth carving provides the best cloud shape prior, while our advection module helps in predicting better cloud shapes with less noise. 
    }
    \label{fig:ablations}
\end{figure*}

\subsection{Quantitative Results}
Table~\ref{tab:quantitative_results} shows our segmentation metrics using the middle upwards-looking view. We see a similar pattern as with the qualitative results, where the neural rendering methods perform the worst, with IBRNet~\cite{wang2021ibrnet} being able to predict cloud coverage better than DVGO~\cite{sun2022dvgo} and MVSNeRF~\cite{chen2021mvsnerf}. 3DeepCT~\cite{3deepct} and VIP-CT~\cite{vip_ct} both perform better, however, we observe that their results only show a marginal improvement when compared against the stand-alone space carving estimate which is used to mask their outputs. Our method, on the other hand, uses our depth carving in a learned manner by utilizing it as an initial coarse volumetric estimate. This results in our method being able to significantly improve on the depth carving estimate.

\subsection{Ablations}
In Figure~\ref{fig:ablations} and Table~\ref{tab:quantitative_ablations} we report ablation results for our model. We specifically consider the following aspects of our method:

\paragraph{Cloud shape guidance strategy.} We show how different cloud shape guidance strategies used for the input to the 3D CNN affect our predicted density fields. Quantitatively, we observe that image features by themselves can estimate cloud coverage to some extent. However, having an initial estimate of the cloud fields is essential for accurate cloud locations. Silhouette-based space carving improves results in some cases but is prone to failure in more complex cases when the carving fails. This can happen due to distant clouds along the horizon or high-altitude cloud layers. Encoding cloud shapes through a truncated signed distance function (TSDF) helps in localizing clouds, but is still susceptible to noise in the depth maps. Our depth carving provides the most accurate cloud location and cloud shape estimate with the least amount of noise.

\begin{table}[h!]
    \centering
    \textbf{Cloud Shape Guidance Strategy}
    \begin{tabularx}{\linewidth}{p{4cm} c c}
        \toprule
        & $\mathcal{J} \uparrow$ & Coverage Error $\downarrow$ \\
        \midrule
        Image Features Only & 50.2 & 15.9 \\
        + Space Carving & 53.7 & 28.7 \\
        + TSDF & 57.5 & 15.3 \\
        + Depth Carving & \textbf{60.9} & \textbf{10.6} \\
        \bottomrule
    \end{tabularx}
    \newline
    \vspace*{-0.25mm}
    \newline
    \textbf{Advection Window Sizes \phantom{y}}
    \begin{tabularx}{\linewidth}{p{4cm} c c}
        \toprule
        & $\mathcal{J} \uparrow$ & Coverage Error $\downarrow$ \\
        \midrule
        Without Advection & 60.9 & 10.6 \\
        + Advection (5 Frames) & 63.1 & 8.1 \\
        + Advection (20 Frames) & \textbf{64.1} & 6.7 \\
        + Advection (40 Frames) & 63.4 &  \textbf{6.3} \\
        \bottomrule
    \end{tabularx}
    \newline
    \vspace*{-0.25mm}
    \newline
    \textbf{Our modules with VIP-CT \phantom{y}}
    \begin{tabularx}{\linewidth}{p{4cm} c c}
        \toprule
        & $\mathcal{J} \uparrow$ & Coverage Error $\downarrow$ \\
        \midrule
        VIP-CT & \textbf{57.3} & 33.0 \\
        + Depth Carving & 52.7 & 13.6 \\
        + Depth C. + Advection & 54.8 & \textbf{11.0} \\
        \bottomrule
    \end{tabularx}
    \caption{\textbf{Ablation study.} We ablate our two main components, the stereo depth carving, and the advection module.}
    \label{tab:quantitative_ablations}
\end{table}

\paragraph{Advection module.} We observe that our advection module outputs reconstructions with smoother cloud shapes that are less noisy. By experimenting with different window sizes, we find there is a balance between having a large enough window size to enforce good temporal consistency and a small enough window size such that wind speeds or cloud shapes do not significantly differ within the same window.

\paragraph{Our modules with VIP-CT.} We find that our depth carving can improve results with VIP-CT~\cite{vip_ct} as can be seen from the improved cloud coverage error. However, unlike our method which refines the depth carving with a 3D CNN, VIP-CT uses it to mask the output in an unlearned manner. This results in limited improvement and can even harm results when the depth maps are incorrect, resulting in a worse Jaccard similarity. Our advection module also improves results for VIP-CT. However, as the pre-advected volumes need to be of good quality, such that a good wind profile can be fit, we find that the improvement is not as substantial as with our method.

\section{Limitations} Despite achieving state-of-the-art results in volumetric reconstruction for clouds in a few-view setting, our framework still has some limitations. Currently, our method only predicts the density fields but could be extended to predict other volume properties, such as droplet size and anisotropic scattering parameters. Finally, our physics-based wind field is modeled under the assumption that cloud motion is predominantly restricted to large-scale translations. While our wind field is a good approximation, clouds have more complex motions that can be investigated in future work.

\section{Conclusion}
We have proposed a framework for obtaining volumetric density estimates from sparse stereo views using a 3D CNN. Our contributions are two-fold. First, we introduce a stereo module that uses depths to carve empty space in front of volumes, giving a coarse volumetric estimate that enables good reconstructions even with limited views. Second, we present an advection module that models the smooth motion of volumes, integrating densities from multiple timesteps into a single refined prediction. To evaluate our proposed framework, we have collected a real-world cloud dataset which we use to demonstrate that our model is able to reconstruct large-scale cloud fields. We believe that our model makes a crucial step towards obtaining a more accurate understanding of our atmosphere.

\vspace{-0.25cm}
\paragraph{Acknowledgements} This work was supported by the Natural Environment Research Council (grant nos. NE/X012255/1, NE/X018539/1), the Royal Society (grant no. URF/R1/191602), and the Portuguese Foundation for Science and Technology (grant no. 2022.12484.BD).

\newpage
{
    \small
    \bibliographystyle{ieeenat_fullname}
    \bibliography{main}
}

\end{document}